\documentclass[conference]{IEEEtran}
\IEEEoverridecommandlockouts
\usepackage{cite}
\usepackage{amsmath,amssymb,amsfonts}
\usepackage{algorithmic}
\usepackage{graphicx}
\usepackage{textcomp}
\usepackage{microtype}

\usepackage[dvipsnames]{xcolor}
\usepackage[colorlinks=true,allcolors=NavyBlue]{hyperref}

\def\BibTeX{{\rm B\kern-.05em{\sc i\kern-.025em b}\kern-.08em
    T\kern-.1667em\lower.7ex\hbox{E}\kern-.125emX}}

\DeclareMathOperator{\tr}{tr}
\DeclareMathOperator{\cov}{cov}
\DeclareMathOperator{\vectorize}{v}

\DeclareMathOperator*{\etr}{etr}

\begin{document}

\title{Depth induces scale-averaging in overparameterized linear Bayesian neural networks
\thanks{This work was supported by the Harvard Data Science Initiative Competitive Research Fund, the Harvard Dean’s Competitive Fund for Promising Scholarship, and a Google Faculty Research Award.}
}

\author{\IEEEauthorblockN{Jacob A. Zavatone-Veth}
\IEEEauthorblockA{\textit{Department of Physics} \\
\textit{Harvard University}\\
Cambridge, MA, United States \\
\texttt{jzavatoneveth@g.harvard.edu}}
\and
\IEEEauthorblockN{Cengiz Pehlevan}
\IEEEauthorblockA{\textit{John A. Paulson School of Engineering and Applied Sciences} \\
\textit{Harvard University}\\
Cambridge, MA, United States \\
\texttt{cpehlevan@seas.harvard.edu}}
}

\maketitle

\begin{abstract}
Inference in deep Bayesian neural networks is only fully understood in the infinite-width limit, where the posterior flexibility afforded by increased depth washes out and the posterior predictive collapses to a shallow Gaussian process. Here, we interpret finite deep linear Bayesian neural networks as data-dependent scale mixtures of Gaussian process predictors across output channels. We leverage this observation to study representation learning in these networks, allowing us to connect limiting results obtained in previous studies within a unified framework. In total, these results advance our analytical understanding of how depth affects inference in a simple class of Bayesian neural networks. 
\end{abstract}

\begin{IEEEkeywords}
Bayesian inference, neural networks, representation learning
\end{IEEEkeywords}

\section{Introduction}

Understanding the effect of depth and width on inference is among the central goals of the modern theory of neural networks. Recent theoretical advances have elucidated the behavior of networks in the infinite-width limit, in which the complexity introduced by depth washes out and inference is described by Gaussian process regression \cite{neal1996priors,williams1997computing,lee2018deep,matthews2018gaussian,yang2019scaling,hron2020exact,lee2020finite}. However, inference at finite widths, where hidden layers retain the flexibility to learn task-relevant representations, remains incompletely understood \cite{aitchison2020bigger,zv2021exact,zv2021asymptotics,lee2020finite,roberts2021principles}. In the setting of gradient-based maximum likelihood optimization, some insights have been gained through the study of finite overparameterized deep linear neural networks \cite{fukumizu1998effect,saxe2013exact,yun2021unifying,atanasov2021kernel}. In the fully Bayesian setting, the behavior of this simple class of models has been characterized in several limiting cases, including asymptotically at large but finite width \cite{neal1996priors,aitchison2020bigger,zv2021asymptotics,li2021statistical,roberts2021principles}. However, a unifying perspective on these results is lacking, and our understanding of inference in deep linear Bayesian neural networks (henceforth $\ell$BNNs) at finite width remains incomplete.  

Here, we make the following contributions toward a more comprehensive understanding of $\ell$BNN inference: 
\begin{enumerate}
    \item 
    We express the moment generating function of the posterior predictive of a finite, overparameterized deep $\ell$BNN as a data-dependent continuous scale mixture of Gaussian process (GP) generating functions. This scale average induces coupling across output channels, and compliments previous interpretations of deep $\ell$BNNs in terms of mixing over an adaptive kernel distribution \cite{pleiss2021limitations,aitchison2021deep}. This observation is mathematically straightforward, but yields some useful insights into inference in finite $\ell$BNNs. We extend this argument to compute the posterior mean feature kernel of  the network's first layer, allowing us to study the representations learned by finite $\ell$BNNs. 
    
    \item 
    We study the asymptotic behavior of these scale mixtures in several limits, allowing us to connect our results to previous work on the asymptotics of $\ell$BNNs \cite{zv2021asymptotics,aitchison2020bigger,roberts2021principles,li2021statistical}. We identify several interesting areas for future investigation, and point to challenges for precise characterization of how $\ell$BNNs behave in certain asymptotic regimes.
    
\end{enumerate}

\section{Setup}\label{sec:setup}

We begin by defining our setup and our notation, which is mostly standard \cite{horn2012matrix,vershynin2018high,magnus2019matrix,williams2006gaussian}. Depending on context, $\Vert \cdot \Vert$ will denote the $\ell_2$ norm on vectors or the Frobenius norm on matrices. We will use the shorthand that integrals without specified domains are taken over all real matrices of the implied dimension. We use the standard Loewner order on real symmetric matrices, such that $A \succeq 0$ (respectively $A \succ 0$) means that the matrix $A$ is positive semi-definite, or PSD (respectively positive-definite, or PD). For a matrix $A \in \mathbb{R}^{p \times n}$, we let $\vectorize(A) \in \mathbb{R}^{pn}$ be its row-major vectorization. Then, denoting the Kronecker product by $\otimes$, we have $\vectorize(ABC) = (A \otimes C^{\top}) \vectorize(B)$ for conformable matrices $A$, $B$, and $C$. For brevity, we define the shorthand $\etr(X) = \exp\tr(X)$.

For a set of compatibly-sized matrices $W_{1} \in \mathbb{R}^{n_{1} \times n_{0}}$, $W_{2} \in \mathbb{R}^{n_{2} \times n_{1}}$, \ldots, $W_{d} \in \mathbb{R}^{n_{d} \times n_{d-1}}$, we define a depth-$d$ $\ell$BNN as the linear map
\begin{equation}
\begin{split}
    f : \mathbb{R}^{n_0} &\to \mathbb{R}^{n_d}
    \\ 
    x &\mapsto W_{d} \cdots W_{1} x.
\end{split}
\end{equation}
We will assume that the ``hidden layer widths" $n_{1},n_{2},\ldots,n_{d-1}$ are all greater than or equal to the output dimension $n_{d}$, such that the rank of the end-to-end weight matrix $W_{d} \cdots W_{1}$ is not constrained by an intermediate bottleneck. We make the standard choice of isotropic Gaussian priors over the weight matrices:
\begin{equation}
    [W_{\ell}]_{ij} \sim_{\textrm{i.i.d.}} \mathcal{N}\left(0,n_{\ell-1}^{-1}\right),
\end{equation}
with variances chosen such that the prior variances of the activations at any layer do not diverge with increasing width \cite{mackay1992practical,neal1996priors,williams1997computing,lee2018deep,matthews2018gaussian,yang2019scaling,hron2020exact}. One could allow general layer-dependent variances $\sigma_{\ell}^{2}/n_{\ell}$, but for $\ell$BNNs the additional factors can always be absorbed into the definition of the input so long as they are finite and non-zero. Thus, for the sake of notational clarity, we make the simplest choice of prior variances. 

For a training dataset $\mathcal{D} = \{(x_{\mu},y_{\mu})\}_{\mu=1}^{p}$ of $p$ examples, we choose an isotropic Gaussian likelihood
\begin{equation} \label{eqn:likelihood}
    p(\mathcal{D}\,|\,W_{1},\ldots,W_{d}) \propto \exp\left(-\frac{\beta}{2} \sum_{\mu=1}^{p} \Vert f(x_{\mu}) - y_{\mu} \Vert^2 \right);
\end{equation}
we will refer to the inverse variance $\beta \geq 0$ as the \emph{inverse temperature} in analogy with statistical mechanics. The Bayes posterior over the weight matrices is then given up to normalization as $p(W_{1},\ldots,W_{d} \,|\, \mathcal{D}) \propto p(\mathcal{D}\,|\,W_{1},\ldots,W_{d}) p(W_{1}) \cdots p(W_{d})$.

We collect the training inputs and targets into data matrices $X \in \mathbb{R}^{p \times n_{0}}$ and $Y \in \mathbb{R}^{p \times n_{d}}$ with elements $X_{\mu j} = x_{\mu,j}$ and $Y_{\mu j} = y_{\mu,j}$, respectively. We will sometimes find it useful to consider a differentiated test dataset $\hat{\mathcal{D}} = \{(\hat{x}_{\hat{\mu}},\hat{y}_{\hat{\mu}})\}_{\hat{\mu}=1}^{\hat{p}}$ with corresponding data matrices $\hat{X} \in \mathbb{R}^{\hat{p} \times n_{0}}$ and $\hat{Y} \in \mathbb{R}^{\hat{p} \times n_{d}}$. For these data, we define the associated normalized Gram matrices $G_{xx} \equiv n_{0}^{-1} X X^{\top}$, $G_{x\hat{x}} \equiv n_{0}^{-1} X \hat{X}^{\top}$, $G_{\hat{x}\hat{x}} \equiv n_{0}^{-1} \hat{X} \hat{X}^{\top}$, $G_{yy} \equiv n_{d}^{-1} YY^{\top}$, and $G_{\hat{y}\hat{y}} \equiv n_{d}^{-1} \hat{Y} \hat{Y}^{\top}$. Our assumptions on the data will be given purely in terms of conditions on these Gram matrices. In particular, we will assume that the training input Gram matrix $G_{xx}$ is invertible; other conditions will be introduced as needed. We note that this invertibility condition, combined with our assumption that the hidden layer widths are wide enough such that the end-to-end weight matrix is not rank-constrained, means that the $\ell$BNNs we consider can linearly interpolate their training data, and are thus overparameterized. 

\section[Scale-averaging in deep LBNNs]{Scale-averaging in deep $\ell$BNNs} \label{sec:scale}

\subsection{The function-space prior as a scale mixture}

We begin with the nearly trivial observation that, for some input data matrix $X$, the induced prior over network outputs $F = X W_{1}^{\top} \cdots W_{d}^{\top}$ can be expressed as a continuous scale mixture of matrix Gaussians. This expression will prove useful in our subsequent study of the posterior predictive by allowing us to compute integrals over network outputs rather than over network weights. This simplification is allowed thanks to the fact that the likelihood \eqref{eqn:likelihood} models the targets $Y$ as being independent of the parameters given the network outputs. 

Recall from \S\ref{sec:setup} that the prior distribution of the first layer's weight matrix is a matrix Gaussian:
\begin{align}
    W_{1}^{\top} \sim \mathcal{MN}_{n_{1} \times n_{0}}(0,n_{0}^{-1} I_{n_{0}},I_{n_{1}}).
\end{align}
Then, for $W_{2},\ldots,W_{d}$ fixed, the distribution of $F$ induced by the prior over $W_{1}$ can be read off using the properties of the matrix Gaussian under linear transformations \cite{vershynin2018high}:
\begin{align}
    F = X W_{1}^{\top} (W_{2}^{\top} \cdots W_{d}^{\top}) \sim \mathcal{MN}_{p \times n_{d}}(0, G_{xx}, L), 
\end{align}
where we have recognized the normalized Gram matrix $G_{xx}$ and defined the $n_{d} \times n_{d}$ matrix 
\begin{equation}
    L \equiv W_{d} \cdots W_{2} W_{2}^{\top} \cdots W_{d}^{\top}.
\end{equation}
For this to make sense, both $G_{xx}$ and $L$ must be of full rank. As stated in \S\ref{sec:setup}, we assume the dataset to be such that $G_{xx}$ is invertible. Moreover, denoting the prior distribution over $L$ induced by the priors over $W_{2},\ldots,W_{d}$ by $\varpi$, the stated assumption that $n_{1},\ldots,n_{d-1} \geq n_{d}$, implies that $L$ is invertible $\varpi$-almost surely \cite{horn2012matrix,vershynin2018high}. 

Using the law of total expectation, we conclude that the prior over outputs for any fixed set of inputs with invertible Gram matrix is a continuous scale mixture of matrix Gaussians, with the prior density explicitly given as
\begin{equation} \label{eqn:prior_density}
\begin{split}
    p_{d}(F \,|\,X) = \mathbb{E}_{L \sim \varpi} \bigg[& (2\pi)^{-n_{d}p/2} \det(L \otimes G_{xx})^{-1/2} \\ &\times \etr\left(-\frac{1}{2} L^{-1} F^{\top} G_{xx}^{-1} F\right) \bigg].
\end{split}
\end{equation}

For a depth-two network, $\varpi$ is a Wishart distribution $L \sim \mathcal{W}_{n_2}(n_{1}^{-1} I_{n_2}, n_1)$, which simplifies to a scalar Gamma-distributed random variable $\lambda \sim \operatorname{Gamma}(n_{1}/2,2/n_{1})$ when $n_{2} = 1$ \cite{vershynin2018high}. These results allow one to easily write down the density of $\varpi$ with respect to Lebesgue measure in the two-layer case. We note that the density of $\varpi$ is expressible for deeper networks in terms of the Meijer $G$-function \cite{dlmf}; we will not further pursue this line of analysis in the present work. One could also integrate out weight matrices beyond $W_{1}$, but this would yield more complicated formulas for the prior density, which do not permit easy analysis of the posterior predictive \cite{zv2021asymptotics,zv2021exact}. In particular, it is unclear how one might obtain an exact expression for the joint function-space prior density for all $p$ examples \cite{zv2021exact}.

\subsection{The cumulant generating function of the posterior predictive}

We now exploit the observations of the previous section to study the posterior predictives of finite $\ell$BNNs. To do so, we will consider the moment generating function 
\begin{equation} \label{eqn:general_genfun}
    Z(\beta,J) = \mathbb{E}_{W_{1},\cdots,W_{d} \,|\,X,Y} \etr(J^{\top} W_{d}\cdots W_{1} \hat{X}^{\top})
\end{equation}
of the posterior predictive for some test data $\hat{X}$. To leverage the mixture-of-Gaussians interpretation of the prior, we express the generating function as an integral over function outputs $F = X W_{1}^{\top} \cdots W_{d}^{\top}$ and $\hat{F} = \hat{X} W_{1}^{\top} \cdots W_{d}^{\top}$, yielding
\begin{equation} \label{eqn:function_space_genfun_integral}
\begin{split}
    Z(\beta,J) \propto \int dF\,d\hat{F}\, & \exp\left(\tr(\hat{F}^{\top} J) - \frac{\beta}{2} \Vert F-Y \Vert^{2} \right) \\& \times p_{d}(F,\hat{F} \,|\,X,\hat{X})
\end{split}
\end{equation}
in terms of the joint prior $p_{d}(F,\hat{F} \,|\,X,\hat{X})$, where the implied constant of proportionality ensures that $Z(\beta,0) = 1$. Here, the joint prior $p_{d}(F,\hat{F} \,|\,X,\hat{X})$ is given by substituting the combined dataset $\begin{bmatrix} X \\ \hat{X} \end{bmatrix}$ into \eqref{eqn:prior_density} under the temporary assumption that the Gram matrix of the combined dataset is invertible. 

We now exchange integration over $F$ and $\hat{F}$ with expectation over the scale matrix $L$, which allows us to evaluate the Gaussian integrals over $F$ and $\hat{F}$ exactly. This calculation is easily performed using row-major vectorization \cite{magnus2019matrix}; we defer a detailed sketch to the Appendix and merely summarize the result here. We define the $p n_{d} \times pn_{d}$ symmetric matrix
\begin{equation} \label{eqn:gammaL}
    \Gamma_{L} \equiv I_{p n_{d}} + \beta G_{xx} \otimes L,
\end{equation}
the $\hat{p} n_{d} \times \hat{p} n_{d}$ symmetric matrix
\begin{equation} \label{eqn:sigmaL}
    \Sigma_{L} \equiv G_{\hat{x}\hat{x}} \otimes L - \beta (G_{x\hat{x}}^{\top} \otimes L) \Gamma_{L}^{-1} (G_{x\hat{x}} \otimes L),
\end{equation}
and the $\hat{p} n_{d}$-dimensional vector
\begin{equation} \label{eqn:muL}
    \mu_{L} \equiv \beta (G_{x\hat{x}}^{\top} \otimes L) \Gamma_{L}^{-1}  \vectorize(Y).
\end{equation}
We let $\rho$ be a probability measure over $n_{d} \times n_{d}$ positive semidefinite matrices, defined by its density
\begin{equation}
    \frac{d\rho}{d\varpi} \propto \det\left(\Gamma_{L} \right)^{-1/2} \exp\left(-\frac{1}{2} \beta \vectorize(Y)^{\top} \Gamma_{L}^{-1} \vectorize(Y) \right)
\end{equation}
with respect to $\varpi$; the implied constant of proportionality ensures that $\int_{L \succeq 0} d\rho(L) = 1$. Then,
\begin{equation} \label{eqn:posterior_genfun}
    Z(\beta,J) = \mathbb{E}_{L \sim \rho} \exp\left(\mu_{L}^{\top} \vectorize(J) + \frac{1}{2} \vectorize(J)^{\top} \Sigma_{L} \vectorize(J) \right).
\end{equation}
From this moment generating function, we can immediately read off that the mean and covariance of the posterior predictive are
\begin{equation}
    \langle \hat{F}_{\hat{\mu} j} \rangle = \mathbb{E}_{L \sim \rho}  \mu_{L}^{\top} \vectorize(\chi_{\hat{\mu} j})
\end{equation}
and
\begin{equation}
\begin{split}
    \cov(\hat{F}_{\hat{\mu} j},\hat{F}_{\hat{\nu} k}) &= \mathbb{E}_{L \sim \rho}  \vectorize(\chi_{\hat{\mu} j})^{\top} \Sigma_{L} \vectorize(\chi_{\hat{\nu} k}) \\&\quad + \cov_{L \sim \rho}\left( \mu_{L}^{\top} \vectorize(\chi_{\hat{\mu} j}  ), \mu_{L}^{\top} \vectorize(\chi_{\hat{\nu} k} )  \right),
\end{split}
\end{equation}
respectively, where we define the $\hat{p} \times n_{2}$ matrix $[\chi_{\hat{\mu} j}]_{\hat{\rho} l} = \delta_{\hat{\mu} \hat{\rho}} \delta_{jl}$. We remark that all of these results extend to the training set predictor with the replacement $G_{x\hat{x}} \gets G_{xx}$. 

We recognize \eqref{eqn:posterior_genfun} as a scale-average of the GP generating function of a single-layer $\ell$BNN, for which $L = I_{n_{d}}$ \cite{neal1996priors,williams1997computing,lee2018deep,matthews2018gaussian,yang2019scaling,hron2020exact,williams2006gaussian}. Similarly, the mean predictor is a scale-average of GP mean predictors, while the predictor covariance includes an additional term beyond the average of the GP covariance, as per the law of total covariance \cite{vershynin2018high}. We emphasize that the scale distribution $\rho$ is data-dependent: depth allows the $\ell$BNN to adaptively couple its output channels in a way that a single-layer network cannot. We finally remark that, unlike in studies of gradient-based maximum likelihood estimation in deep linear networks \cite{fukumizu1998effect,saxe2013exact,yun2021unifying,atanasov2021kernel}, no exceptional assumptions on the weight distribution or data are required to obtain this intuitive picture.

The above results are rendered somewhat complicated by the need to average over $n_{d} \times n_{d}$ PSD matrices. If $n_{d} = 1$, the situation simplifies substantially, as the scale variable is now a scalar $\lambda$, and the Kronecker products can be eliminated. Concretely, for $\lambda \geq 0$, we define
\begin{align}
    \Gamma_{\lambda} &\equiv I_{p} + \beta \lambda G_{xx} \in \mathbb{R}^{p \times p},
    \\
    \Sigma_{\lambda} &\equiv \lambda G_{\hat{x}\hat{x}} - \beta \lambda^{2} G_{x\hat{x}}^{\top} \Gamma_{\lambda}^{-1} G_{x\hat{x}} \in \mathbb{R}^{\hat{p} \times \hat{p}}, \quad \textrm{and}
    \\
    \mu_{\lambda} &\equiv \lambda \beta  G_{x\hat{x}}^{\top} \Gamma_{\lambda}^{-1} y \in \mathbb{R}^{\hat{p}},
\end{align}
and let $\rho$ be a probability measure on $[0,\infty)$, defined by its density
\begin{equation}
    \frac{d\rho}{d\varpi} \propto \det(\Gamma_{\lambda})^{-1/2} \exp\left(-\frac{1}{2} \beta y^{\top} \Gamma_{\lambda}^{-1} y\right)
\end{equation}
with respect to $\varpi$; the implied constant of proportionality ensures that $\int_{0}^{\infty} d\rho(\lambda) = 1$. Then, the cumulant generating function of the posterior predictive of a deep $\ell$BNN with scalar output can be expressed as
\begin{equation}
    Z(\beta,j) = \mathbb{E}_{\lambda \sim \rho} \exp\left(\mu_{\lambda}^{\top} j + \frac{1}{2} j^{\top} \Sigma_{\lambda} j\right).
\end{equation}
From this, we obtain correspondingly simplified expressions for the predictor mean and covariance, which reduce to 
\begin{equation}
    \langle \hat{f} \rangle = \mathbb{E}_{\lambda \sim \rho} \mu_{\lambda}
\end{equation}
and
\begin{equation}
    \cov(\hat{f}) = \mathbb{E}_{\lambda \sim \rho} \Sigma_{\lambda} + \cov_{\lambda \sim \rho}( \mu_{\lambda} ),
\end{equation}
respectively. Again, these results represent scale-averages of shallow GP predictors, but they are of a simpler form thanks to the lack of mixing between outputs. Even in this simplified setting, and even if one makes a further restriction to the case in which there is only a single training example, the averages defy exact analysis for general values of the hyperparameters due to the terms of the form $1/(1+\beta G_{xx} \lambda)$ in the exponent \cite{dlmf}.

\subsection[The zero-temperature limit]{The zero-temperature limit} \label{sec:lowtemp}

Though analysis of the scale distribution is challenging for general values of the likelihood variance, the situation simplifies somewhat in the zero-temperature limit $\beta \to \infty$ of vanishing likelihood variance. In this limit, the likelihood tends to a collection of Dirac masses that enforce the constraint that the $\ell$BNN interpolates its training set. For this interpretation to be sensible at the level of the posterior predictive, the training dataset must be linearly interpolatable, i.e., there must exist some matrix $W \in \mathbb{R}^{n_{0} \times n_{d}}$ such that $XW=Y$. We will focus on this case, and operate under the assumption that the training dataset Gram matrix $G_{xx}$ is invertible. Then, we expect all expectations over $L$ to be sufficiently regular such that we can interchange the limit in $\beta$ with the integrals, which should allow us to compute them using the pointwise limit of the density $d\rho/d\varpi$. We note that, though this limit is convenient for theoretical analysis, it is somewhat unnatural from a Bayesian perspective, as it models the targets as a deterministic function of the outputs \cite{mackay1992practical,wilson2020bayesian}. 

Under these regularity assumptions, we expect to have the almost-sure low-temperature limit $\beta \Gamma_{L}^{-1} \to G_{xx}^{-1} \otimes L^{-1}$, which yields the almost-sure limiting behavior
\begin{align}
    \mu_{L} &\to (G_{x\hat{x}}^{\top} G_{xx}^{-1} \otimes I_{n_2}) \vectorize(Y),
    \\
    \Sigma_{L} &\to (G_{\hat{x}\hat{x}} - G_{x\hat{x}}^{\top} G_{xx}^{-1} G_{x\hat{x}}) \otimes L.
\end{align}
Then, the limiting mean predictor simplifies to
\begin{equation}
    \lim_{\beta \to \infty} \langle \hat{F} \rangle = G_{x\hat{x}}^{\top} G_{xx}^{-1} Y.
\end{equation}
This precisely corresponds to the least-norm pseudoinverse solution to the system $XW=Y$, which is intuitively sensible. Moreover, we have the limiting covariance
\begin{equation}
\begin{split}
    \lim_{\beta \to \infty} \cov(\hat{F}_{\hat{\mu} j},\hat{F}_{\hat{\nu} k}) &= (G_{\hat{x}\hat{x}} - G_{x\hat{x}}^{\top} G_{xx}^{-1} G_{x\hat{x}})_{\hat{\mu} \hat{\nu}} \\&\quad \times  \lim_{\beta \to \infty} \mathbb{E}_{L \sim \rho} L_{jk},
\end{split}
\end{equation}
which is precisely the GP posterior sample-sample covariance, multiplied by a coupling between output channels. 

This argument also yields an approximate density 
\begin{equation}
    \frac{d\rho}{d\varpi} \propto \frac{1}{\det(L)^{p/2}} \etr\left(-\frac{1}{2} Y^{\top} G_{xx}^{-1} Y L^{-1} \right).
\end{equation}
In the two-layer case, where $L$ follows a Wishart distribution, the limiting density of $\rho$ with respect to Lebesgue measure on PSD matrices should then be given as
\begin{equation} \label{eqn:lowtemp_density}
\begin{split}
    \frac{d\rho}{dL} &\propto \det(L)^{(n_{1}-p)/2-(n_{2}+1)/2} \\&\quad\times \etr\left(-\frac{1}{2} (n_{1} L + Y^{\top} G_{xx}^{-1} Y L^{-1}) \right).
\end{split}
\end{equation}
This implies that $L$ follows a matrix generalized inverse Gaussian (MGIG) distribution at low temperatures \cite{butler1998generalized,fazayeli2016matrix}:
\begin{equation}
    L \sim \mathcal{MGIG}_{n_{2}}\left( Y^{\top} G_{xx}^{-1} Y, n_{1} I_{n_2}, \frac{n_{1}-p}{2} \right).
\end{equation}
This observation yields several insights. First, it implies that the moment generating functions of $L$ and $L^{-1}$ are given in terms of Bessel functions of matrix argument of the second kind $B_{\nu}(Z)$ \cite{butler1998generalized,fazayeli2016matrix,butler2003laplace,herz1955bessel,dlmf}. Second, neither the mean $\mathbb{E} L$ nor the reciprocal mean $\mathbb{E} L^{-1}$ of the MGIG are known in closed form for general values of the parameters \cite{butler1998generalized,fazayeli2016matrix}. We will therefore resort to studying the behavior of these expectations in various asymptotic limits in \S\ref{sec:asymptotics}. However, reasonably efficient algorithms for sampling from the MGIG are available; the situation is of course particularly simple when $n_{2}=1$ \cite{fazayeli2016matrix}. Therefore, this formulation could allow faster numerical studies of two-layer $\ell$BNNs at low temperatures than is possible through na{\"i}ve sampling of the weights, as the dimensionality of the search space is reduced from $n_{0}n_{1}+n_{1}n_{2}$ to $n_{2}^{2}$.

\section[Average first-layer feature kernels in LBNNs]{Average first-layer feature kernels in $\ell$BNNs} \label{sec:kernel}

We now use the methods of \S\ref{sec:scale} to study the average feature kernels of deep $\ell$BNNs. For technical convenience, we restrict our attention to the kernel of the first hidden layer evaluated on the training set:
\begin{equation}
    K \equiv \frac{1}{n_{1}} X W_{1} W_{1}^{\top} X^{\top}.
\end{equation}
Then, we can proceed as before to integrate $W_{1}$ out of the posterior moment generating function of $K$:
\begin{equation}
    Z(\beta,J) = \mathbb{E}_{W_{1},\ldots,W_{d}\,|\,X,Y} \etr\left( \frac{1}{n_1} J X W_{1}^{\top} W_{1} X^{\top} \right).
\end{equation}
As discussed in the Appendix, the required computation is straightforward as all integrals are Gaussian. Whereas we considered the full generating function of the posterior predictive, we focus only on the posterior mean of the kernel. Defining the $p \times p$ scale-dependent matrix
\begin{equation}
\begin{split}
    [\Delta_{L}]_{\mu\nu} &\equiv \beta^{2} \vectorize(Y)^{\top} \Gamma_{L}^{-1} (G_{xx} \chi_{\mu\nu} G_{xx} \otimes L ) \Gamma_{L}^{-1} \vectorize(Y) \\&\quad - \beta \tr[\Gamma_{L}^{-1} (G_{xx} \chi_{\mu\nu} G_{xx} \otimes L)]
\end{split}
\end{equation}
for $\chi_{\mu\nu}$ is the $p\times p$ matrix $[\chi_{\mu\nu}]_{\rho\lambda} = \delta_{\mu\rho}\delta_{\nu\lambda}$, the posterior-averaged feature kernel can be expressed as
\begin{equation} \label{eqn:featurekernel}
    \langle K \rangle = G_{xx} + \frac{1}{n_{1}} \mathbb{E}_{L \sim \rho} \Delta_{L}.
\end{equation}
As the shallow GP result is simply $\langle K \rangle = G_{xx}$, this yields a natural interpretation of the mean kernel of a finite-width deep $\ell$BNN as the GP kernel plus some correction. Though the complexity of the matrix $\Delta_{L}$ renders this result somewhat less than fully transparent, the situation again simplifies for the case of scalar output, for which we have
\begin{equation}
    \Delta_{\lambda} = \lambda \beta^2 G_{xx} \Gamma_{\lambda}^{-1}  y y^{\top} \Gamma_{\lambda}^{-1} G_{xx} - \lambda\beta G_{xx} \Gamma_{\lambda}^{-1} G_{xx}.
\end{equation}
We observe that the first term in this result is the outer product of the non-scale-averaged mean training set predictor $\beta G_{xx} \Gamma_{\lambda}^{-1}  y$ with itself. Strikingly, the matrix $\Delta_{\lambda}$---when evaluated at $\lambda = 1$---is precisely the matrix that appears as the asymptotic correction to the average kernel computed in our previous work \cite{zv2021asymptotics}. 

Following the discussion of \S\ref{sec:lowtemp}, we have the almost-sure pointwise low-temperature limit
\begin{equation}
    \lim_{\beta \to \infty} \Delta_{L} = Y L^{-1} Y^{\top} - n_{d} G_{xx}.
\end{equation}
Thus, to compute the average kernel at low temperatures, we must compute the limiting reciprocal mean of $L$. As noted in \S\ref{sec:lowtemp}, this is not known in closed form.

\section[Asymptotics of LBNNs]{Asymptotic behavior of $\ell$BNNs} \label{sec:asymptotics}

We now consider the asymptotic behavior of $\ell$BNNs in various limits, allowing us to connect our results to those of previous works. So as to make contact with as many previous works as possible \cite{neal1996priors,williams1997computing,lee2018deep,matthews2018gaussian,yang2019scaling,hron2020exact,zv2021asymptotics,roberts2021principles,li2021statistical,aitchison2020bigger}, we will largely focus on the behavior of the average kernel $\langle K \rangle$. In all cases, we will assume that the hidden layer widths $n_{1},\ldots,n_{d-1}$ are of a comparable  scale $n$, such that the ratios $n_{\ell}/n$ remain fixed as $n$ is taken to be large. As in the rest of the paper, we will assume that $G_{xx}$ is invertible, which requires that $n_{0} \geq p$. Thus, in limits in which $p$ is taken to be large, we implicitly also take $n_{0}$ to be large. We will only consider limits in which the depth is held fixed and finite, or, at least, tends to infinity far more slowly than the hidden layer width, such that $d/n$ is perturbatively small \cite{zv2021asymptotics,roberts2021principles,hanin2021random}. For the sake of analytical tractability, will often restrict our attention to two-layer networks ($d=2$) in the zero-temperature limit ($\beta \to \infty$). For notational brevity, we define the ratios $\alpha \equiv p/n$ and $\gamma \equiv n_{d}/n$, which, under our assumptions, are bounded as $0 \leq \alpha,\gamma \leq 1$.

\subsection[Infinite width, fixed dataset size and output dimension]{$n \to \infty$, $p$ and $n_{d}$ fixed}

We first consider the regime in which the hidden layer widths tend to infinity with fixed input dimension, output dimension, training dataset size, and depth. This is the most commonly considered asymptotic regime for BNNs \cite{neal1996priors,williams1997computing,lee2018deep,matthews2018gaussian,yang2019scaling,hron2020exact,zv2021asymptotics,roberts2021principles}. In this limit, a simple saddle-point argument shows that the data-dependence in $\rho$ can be neglected, and that the expectations over $L$ should be dominated by the mode of $\varpi$, which is $L_{\ast} = \mathbb{E}_{L \sim \varpi} L = I_{n_d}$ \cite{shun1995laplace}. Applying this result to evaluate the expectations in the posterior predictive \eqref{eqn:posterior_genfun}, we recover the expected correspondence between infinitely-wide BNNs and Gaussian processes \cite{neal1996priors,williams1997computing,lee2018deep,matthews2018gaussian,yang2019scaling,hron2020exact,zv2021asymptotics,roberts2021principles}. 

Moreover, we can use this simple argument to recover the leading asymptotic correction to the average hidden layer kernel computed in our previous work \cite{zv2021asymptotics}. As the expectation in \eqref{eqn:featurekernel} carries an overall factor of $1/n_{1}$, the leading correction is simply given by evaluating $\Delta_{L}$ at the saddle-point value of $L$; corrections to the saddle point at large but finite widths will lead subleading corrections to the kernel \cite{shun1995laplace}. After some algebraic simplification, this yields
\begin{equation}
    \langle K \rangle = G_{xx} + \gamma  G_{xx} \Gamma_{\infty}^{-1} (G_{yy} - \Gamma_{\infty}) \Gamma_{\infty}^{-1} G_{xx} + \mathcal{O}(\gamma^{-2}),
\end{equation}
where $\Gamma_{\infty} \equiv G_{xx} + I_{p}/\beta$. This matches the result of \cite{zv2021asymptotics}, and is consistent with the observation in \S\ref{sec:kernel} that the finite-width kernel in the scalar output setting is simply the average of the asymptotic correction over scales. More generally, one could treat $L$ as a small perturbation of the identity, and use perturbative methods similar to those of our previous work to recover the results on corrections to predictor statistics given there \cite{zv2021asymptotics}. 

\subsection[Infinite dataset size, fixed width and output dimension]{$p \to \infty$, $n$ and $n_{d}$ fixed}

We next consider the regime in which the dataset size is taken to be large relative to the hidden layer width and output dimension. This regime is of interest because one expects posterior concentration to occur in the large-dataset regime \cite{mackay1992practical,wilson2020bayesian}. Focusing on the zero-temperature limit, we make the simple approximation of neglecting all terms in the density \eqref{eqn:lowtemp_density} that do not scale with $p$, leaving $d\rho/dL \propto \exp\{-p[\tr(Y^{\top} G_{xx}^{-1} Y L^{-1})/p + \log\det(L)]/2\}$. We then evaluate the integral over $L$ by a saddle-point approximation, yielding $L = Y^{\top} G_{xx}^{-1} Y/p$. This yields an average kernel of
\begin{equation}
    \langle K \rangle \approx (1-\gamma) G_{xx} + \frac{1}{n_{1}} Y  \left( \frac{1}{p} Y^{\top} G_{xx}^{-1} Y \right)^{-1} Y^{\top}
\end{equation}
under the reasonable assumption that $Y^{\top} G_{xx}^{-1} Y$ is of full rank in this regime. Notably, the correction to the GP kernel need not be vanishingly small.

\subsection[Infinite width and dataset size, fixed output dimension]{$n,p \to \infty$ and $n_{d}$ fixed}

We now consider the limit in which the hidden layer width and training dataset size tend to infinity for fixed depth and output dimension, as previously studied by Li \& Sompolinsky \cite{li2021statistical}. We focus---as those authors did---on the zero-temperature limit, and restrict our attention to the two-layer case for the sake of analytical tractability. Then, exploiting the results of \S\ref{sec:lowtemp}, we expect the expectations over $L$ to be dominated by the mode of the MGIG. Concretely, we neglect terms of order $n_{d}/n$, while keeping the term $Y^{\top} G_{xx}^{-1} Y$ as we expect it to be of order $p$. Then, the mode of $\rho$ is determined by a continuous algebraic Ricatti equation (CARE) \cite{butler2003laplace,fazayeli2016matrix}:
\begin{equation}
    I_{n_2} - L^{-1} \left(\frac{Y^{\top} G_{xx}^{-1} Y}{n_1} \right) L^{-1} - (1 - \alpha) L^{-1}  = 0.
\end{equation}
Using the fact that the solutions of this equation commute with the matrix $Y^{\top} G_{xx}^{-1} Y$ \cite{butler2003laplace}, this is identical to the defining equation for the ``renormalization matrix" of \cite{li2021statistical} in the depth-two case. In particular, using the result of \S\ref{sec:lowtemp} and \S\ref{sec:kernel}, this immediately implies that we recover their results for the predictor statistics and zero-temperature kernel.

After some simplification, the solution to this CARE yields
\begin{equation}
    \langle K \rangle \approx \frac{1}{2} G_{xx} \left[(1 + \alpha) I_{p} + \left( (1-\alpha)^2 I_{p} + 4 \gamma G_{xx}^{-1} G_{yy}\right)^{1/2} \right],
\end{equation}
but there may be corrections to the saddle point at non-vanishing $\gamma$ (in particular, if $n_{d}^{2}$ grows faster than roughly $n^{1/3}$ \cite{shun1995laplace}). To leading order in $\gamma$, we have
\begin{equation}
    \langle K \rangle \approx (1-\gamma) G_{xx} + \frac{\gamma}{1-\alpha} G_{yy} + \mathcal{O}(\gamma^2),
\end{equation}
which is easily seen to agree with the result in the finite-$p$ regime upon expanding when $\alpha \ll 1$. 

\subsection[Infinite width and output dimension, fixed dataset size]{$n,n_{d} \to \infty$ and $p$ fixed}

Our analysis in the preceding sections was facilitated by the fact that the dimensionality of the scale integral remained finite. However, regimes in which the number of outputs tends to infinity with the hidden layer width can also be of interest. In particular, this limit is relevant to the study of autoencoding in high dimensions, and potentially also to classification tasks with many groups (e.g., ImageNet \cite{russakovsky2015imagenet}). Though the same techniques that permit easy asymptotic analysis of other limits cannot be directly applied \cite{shun1995laplace}, the problem of computing kernel statistics can be reformulated as an integral over the $p \times p$ kernel matrices themselves, as noted by Aitchison \cite{aitchison2020bigger}. Then, provided that $p$ is held fixed, the kernel can be computed using a saddle-point approximation. As in the case above, it is easiest to make analytical progress in two-layer networks at zero temperature. There, one finds that the limiting kernel is determined by the solution to the CARE \cite{aitchison2020bigger,aitchison2021solvers} 
\begin{equation}
    G_{xx}^{-1} - \gamma K^{-1} G_{yy} K^{-1} + (\gamma-1) K^{-1} = 0.
\end{equation}
The solution to this CARE yields
\begin{equation}
    \langle K \rangle \approx \frac{1}{2} G_{xx} \left[ (1 - \gamma) I_{p} + \left((1 - \gamma)^2 I_{p} + 4\gamma G_{xx}^{-1} G_{yy} \right)^{1/2} \right] .
\end{equation}
In particular, when $\gamma = 1$, we recover Aichison \cite{aitchison2020bigger}'s result that $K = G_{xx} (G_{xx}^{-1} G_{yy})^{1/2} = (G_{yy} G_{xx}^{-1})^{1/2} G_{xx}$. More generally, we observe that this result is suggestively similar to the kernel in the case of large $p$ and finite $n_{d}$. In particular, this result can be recovered by making what is in principle an unjustified na\"ive Laplace approximation to the integral over $L$ as in the preceding section while keeping terms of order $\gamma$ and ignoring possible corrections to the saddle point from the high-dimensional measure. Further exploration of this will be an interesting subject for future investigation. 

\subsection[Infinite width, output dimension, and dataset size]{$n,n_{d},p\to\infty$}

Finally, one might consider the regime in which the hidden layer width, output dimension, and dataset size tend jointly to infinity. This regime is more challenging to study than those discussed previously, as there is not a clear way to reduce the problem to a finite-dimensional integral. The natural setup for this joint asymptotic limit is a random design teacher-student setting, in which the input examples are independent and identically distributed samples from some distribution and the targets are generated by a linear model with random coefficient matrix. Then, the $\ell$BNN problem is closely related to the random-design linear-rank matrix inference task, which is known to be challenging to analyze \cite{bun2016rotational,barbier2021statistical,maillard2021perturbative}. We direct the interested reader to recent works by Barbier and Macris \cite{barbier2021statistical} and by Maillard \emph{et al.} \cite{maillard2021perturbative} on this problem, and defer more detailed analysis to future work.

\section{Discussion and conclusions}

In this short paper, we have studied some aspects of inference in finite overparameterized $\ell$BNNs. We presented a simple argument that leads to a clear conceptual picture of the effect of depth, and exploited those methods to connect the results of previous studies. However, we note that our approach is specialized to linear networks, and would not extend easily to nonlinear BNNs. Taken together, our results provide some insight into finite-width effects in a model where depth does not affect the hypothesis class, but does affect inference.

The output-mixing scale average interpretation studied in this work compliments previous interpretations of deep BNNs as mixtures of GPs across a data-adaptive distribution of the kernel that measures similarities between input examples. This interpretation has been pursued in a series of recent works by Aitchison and colleagues \cite{aitchison2021deep,aitchison2020bigger,aitchison2021solvers}, starting with the abovementioned work on kernel statistics in deep $\ell$BNNs \cite{aitchison2020bigger}. Those authors have also studied a class of models that generalizes this interpretation of deep BNNs by explicitly fixing prior distributions over data-adaptive kernels, resulting in model predictions \cite{aitchison2020bigger,aitchison2021solvers}. This adaptive-kernel description has also recently been considered by Pleiss and Cunningham \cite{pleiss2021limitations}, who showed that the mean predictor of a two-layer deep GP can be interpreted as a data-dependent mixture of function bases. Their result covers a much broader model class than just $\ell$BNNs---the class of all possible BNNs, linear or nonlinear, is a degenerate subclass of the set of deep GPs---but does not capture higher moments of the posterior predictive. 

For a deep $\ell$BNNs, the adaptive-kernel interpretation arises naturally if one integrates the readout weight matrix $W_{d}$ out of the prior, rather than integrating out the first layer weight matrix $W_{1}$ as we did here \cite{aitchison2021deep,aitchison2020bigger}. Other than in the limit of large output dimension, integrating out $W_{1}$ rather than $W_{d}$ affords some advantages---some merely aesthetic, others technical---if one has the specific objective of analytically characterizing inference in $\ell$BNNs. Both approaches allow for the study of the limit of large width and fixed dataset size, but the need to average over the dataset-size-dimensional kernel matrix makes the large-dataset limit harder to study in the adaptive-kernel interpretation. If one studies the posterior predictive generating function \eqref{eqn:general_genfun} using the adaptive kernel interpretation, one must contend with the need to average over the kernel matrix for the combined train-test set. The blocks of this combined kernel matrix do not appear on equal footing in the generating function because the likelihood only involves the training set; this results in conceptually more complex expressions. Finally, the approach taken here has the advantage of simplifying dramatically for single-output networks; such a simplification is not as obvious in the adaptive-kernel interpretation. 

In concurrent work, Lee \emph{et al.} \cite{lee2021scale} have proposed to manually introduce scale mixing to wide BNNs by fixing priors over the prior variances of the last layer's weights. With this setup, taking the limit of infinite hidden layer width results in a scale mixture of GP predictors. Here, we observe that such scale-averaging arises naturally as an effect of depth in finite-width $\ell$BNNs, hence their setup could be interpreted as manually compensating for the effective loss of depth in an infinite BNN. Based on numerical experiments, they claim that this method can in some cases improve generalization performance relative to that of the fixed-scale GP predictor corresponding to the infinite-width limit of a BNN with fixed prior weight variances. However, importantly, their setup does not consider coupling across multiple output channels. Comprehensive investigation of when data-adaptive scale mixing yields better generalization performance than a fixed-scale GP will be an interesting subject for future investigation. 

To conclude, the results of this work illustrate several important conceptual points. Notably, the behavior of networks with many outputs is qualitatively distinct from those with scalar outputs, as there are interactions between output channels which are apparent neither in the scalar output case nor in the limit of infinite width and fixed output dimension. These interactions render both finite-size and asymptotic analyses more challenging. This issue is not merely one of abstract theoretical interest. Rather, it is potentially relevant to attempts to explain empirical results in deep learning. As modern image recognition tasks often include thousands of classes, the ratio of depth to output dimension of realistic networks may non-negligible \cite{russakovsky2015imagenet}. Thus, we believe that careful analysis of representation learning in joint limits of infinite hidden layer width, output dimension, depth, and dataset size will be an important subject for future work.

\section*{Acknowledgments}

We thank A. Atanasov and B. Bordelon for useful conversations and helpful comments on our manuscript. 

\bibliographystyle{IEEEtran}
\bibliography{IEEEabrv,refs}

\appendix

\section{Derivation of the posterior predictive generating function}\label{app:predictor_genfun}

In this short appendix, we sketch the derivations of our results for the moment generating function of the posterior predictive (reported in \S\ref{sec:scale}) and the posterior average kernel (reported in \S\ref{sec:kernel}). Following the setup in \S\ref{sec:scale}, computation of the moment generating function of the posterior predictive requires only the evaluation of a single Gaussian integral, hence we will omit many intermediate steps for brevity. We proceed under the assumption that the combined Gram matrix
\begin{equation}
    \tilde{G}_{xx} = \begin{bmatrix} G_{xx} & G_{x\hat{x}} \\ G_{x\hat{x}}^{\top} & G_{\hat{x}\hat{x}} \end{bmatrix}
\end{equation}
is invertible; the result extends to the general case by continuity \cite{vershynin2018high,horn2012matrix}. 

We start by using the representation of the generating function as an integral over predictions \eqref{eqn:function_space_genfun_integral} and the expression for the function-space prior density as a continuous scale mixture \eqref{eqn:prior_density}. Then, assuming that that we can apply Fubini's theorem to interchange the integrals over $F$ and $\hat{F}$ with the integral over $L$, our first task is to evaluate the matrix Gaussian integral
\begin{equation}
\begin{split}
    & (2\pi)^{-n_{d}\tilde{p}/2} \det(L)^{-\tilde{p}/2} \det(\tilde{G}_{xx})^{-n_{d}/2} \\&\times \int d\tilde{F}\, \etr\bigg(-\frac{1}{2} \beta (F-Y) (F-Y)^{\top} + \hat{F}^{\top} J \\&\qquad\qquad\qquad -\frac{1}{2} L^{-1} \tilde{F}^{\top} \tilde{G}_{xx}^{-1} \tilde{F} \bigg),
\end{split}
\end{equation}
where $\tilde{F} \equiv [F^{\top}, \hat{F}^{\top}]^{\top}$. This integral is easiest to evaluate using row-major vectorization, for which $\vectorize(\tilde{F}) = [\vectorize(F)^{\top},\vectorize(\hat{F})^{\top}]^{\top}$. Then, defining the matrix
\begin{equation}
    A = \begin{bmatrix} \beta I_{p n_{d}} & 0 \\ 0 & 0 \end{bmatrix} + \tilde{G}_{xx}^{-1} \otimes L^{-1}
\end{equation}
and the vector
\begin{equation}
    b = \begin{bmatrix} \beta \vectorize(Y) \\ \vectorize(J) \end{bmatrix},
\end{equation}
the integral of interest can be expressed as
\begin{equation}
\begin{split}
    & (2\pi)^{-n_{d}\tilde{p}/2} \det(\tilde{G}_{xx} \otimes L)^{-1/2} \\&\times \int d\vectorize(\tilde{F})\,\exp\bigg(-\frac{1}{2} \vectorize(\tilde{F})^{\top} A \vectorize(\tilde{F}) + b^{\top} \vectorize(\tilde{F}) \bigg)
    \\
    &=  \det(\tilde{G}_{xx} \otimes L)^{-1/2} \det(A)^{-1/2} \exp\left(\frac{1}{2} b^{\top} A^{-1} b\right)
\end{split}
\end{equation}
up to a normalizing factor of $\etr(-\beta YY^{\top}/2)$. Using properties of the Kroenecker product, we find after a bit of algebra that \cite{horn2012matrix}
\begin{equation}
    \det(\tilde{G}_{xx} \otimes L) \det(A) = \det[(\tilde{G}_{xx} \otimes L) A ] = \det(\Gamma_{L})
\end{equation}
and
\begin{equation}
\begin{split}
    \frac{1}{2} b^{\top} A^{-1} b - \frac{1}{2} \beta \tr(YY^{\top})
    &= - \frac{1}{2} \beta \vectorize(Y)^{\top} \Gamma_{L}^{-1} \vectorize(Y) \nonumber\\&\quad + \mu_{L}^{\top} \vectorize(J) \nonumber\\&\quad + \frac{1}{2} \vectorize(J)^{\top} \Sigma_{L} \vectorize(J),
\end{split}
\end{equation}
where we have defined the matrices $\Gamma_{L}$ and $\Sigma_{L}$ and the vector $\mu_{L}$ as in \eqref{eqn:gammaL}, \eqref{eqn:sigmaL}, and \eqref{eqn:muL} of the main text, respectively. We then conclude the desired result upon grouping $L$-dependent terms that do not depend on the source $J$ into the density $d\rho/d\varpi$.

The kernel statistics may be derived through an analogous procedure. We start with the posterior moment generating function
\begin{equation}
    Z(\beta,J) = \mathbb{E}_{W_{1},\ldots,W_{d}\,|\,X,Y} \etr\left( - \frac{1}{2} \frac{1}{n_1} J X W_{1}^{\top} W_{1} X^{\top} \right), 
\end{equation}
where the source term is defined with a factor of $-1/2$ for convenience. As before, the first layer weight matrix can be integrated out, yielding 
\begin{equation}
\begin{split}
    & Z(\beta,J) 
    \\&\quad \propto \mathbb{E}_{L \sim \varpi} (2\pi)^{-n_{d}p/2} \det(L)^{-p/2} \det(G_{xx})^{-n_{d}/2} \\&\qquad \times  \int dF\, \exp\left(-\frac{\beta}{2} \Vert F-Y\Vert_{F}^{2}\right) \\&\qquad\qquad\times \etr\left( -\frac{1}{2} L^{-1} F^{\top} G_{xx}^{-1}  \left(I_{p} + \frac{1}{n_{1}} G_{xx} J \right) F\right).
\end{split}
\end{equation}
This is again a Gaussian integral, hence it can be evaluated by direct computation using the vectorization method discussed above. After varying the result with respect to $J$, one obtains the formula \eqref{eqn:featurekernel} reported in the main text.

\end{document}